\documentclass[conference]{IEEEtran}
\IEEEoverridecommandlockouts
\usepackage{cite}
\usepackage{balance}
\usepackage{amsmath,amssymb,amsfonts}
\usepackage{algorithmic}
\usepackage{graphicx}
\usepackage{textcomp}
\usepackage{xcolor}
\usepackage{makecell}

\def\BibTeX{{\rm B\kern-.05em{\sc i\kern-.025em b}\kern-.08em
    T\kern-.1667em\lower.7ex\hbox{E}\kern-.125emX}}
\begin{document}

\title{ML-Based Analysis to Identify Speech Features \\Relevant in Predicting Alzheimer’s Disease
}

\makeatletter
\newcommand{\linebreakand}{%
  \end{@IEEEauthorhalign}
  \hfill\mbox{}\par
  \mbox{}\hfill\begin{@IEEEauthorhalign}
}
\makeatother

\author{

\IEEEauthorblockN{ Yash Kumar\textsuperscript{1}, Piyush Maheshwari\textsuperscript{1}, Shreyansh Joshi\textsuperscript{1}, Veeky Baths\textsuperscript{2}}
\IEEEauthorblockA{
\textit{\textsuperscript{1}Department of Computer Science \& Information Systems}\\
\textit{\textsuperscript{2}Cognitive Neuroscience Lab, Department of Biological Sciences} \\
\textit{BITS Pilani K K Birla Goa Campus, Goa, India}\\
\textsuperscript{1} \{yashkumar1999, piyushm11111, shreyanshjoshi13\}@gmail.com\\
\textsuperscript{2} veeky@goa.bits-pilani.ac.in}
}

\maketitle

\begin{abstract}
Alzheimer’s disease (AD) is a neurodegenerative disease that affects nearly 50 million individuals across the globe and is one of the leading causes of deaths globally. It is projected that by 2050, the number of people affected by the disease would more than double. Consequently, the growing advancements in technology beg the question, can technology be used to predict Alzheimer’s for a better and early diagnosis? In this paper, we focus on this very problem. Specifically, we have trained both ML models and neural networks to predict and classify participants based on their speech patterns. We computed a number of linguistic variables using DementiaBank’s Pitt Corpus, a database consisting of transcripts of interviews with subjects suffering from multiple neurodegenerative diseases. We then trained both binary classifiers, as well as multiclass classifiers to distinguish AD from normal aging and other neurodegenerative diseases. We also worked on establishing the link between specific speech factors that can help determine the onset of AD. Confusion matrices and feature importance graphs have been plotted model-wise to compare the performances of our models. In both multiclass and binary classification, neural networks were found to outperform the other models with a testing accuracy of 76.44\% and 92.05\% respectively. It was also concluded that ‘\%\_PRESP’ (present participle), ‘\%\_3S’ (3rd person present tense markers) were two of the most important speech features for our classifiers in predicting AD.
\end{abstract}

% \begin{IEEEkeywords}
% \end{IEEEkeywords}

\section{Introduction}

Alzheimer's disease is by far the most common type of dementia, accounting for about 70\% of all dementia cases, worldwide. It is a progressive and irreversible neurodegenerative disease that causes brain cells to degenerate and die, thereby debasing the mental faculties of the patient. AD has no ubiquitous definition and is often associated with symptoms like memory loss, language deterioration, mood changes, impaired judgment, and loss of initiative \cite{c:20}.

 Since it is a progressive disease, the symptoms tend to get worse and the patient's health declines year after year. Typically, AD is detected in people over 65 years of age, although the onset begins much before the disease is actually detected. This is primarily because of 2 reasons: patients wait for too long before approaching a  doctor, and secondly, detection of AD depends on the ability of patients to explain their symptoms. Also, many a times it can be difficult for the doctors to diagnose the disease in its early stages because the symptoms are milder and hence can easily be confused with symptoms of other commonly occurring illnesses. As of today, AD has no cure. Existing treatments primarily aim to alleviate the symptoms temporarily and delay its progression. Given it's insidious nature, the motive behind this work of ours was to leverage the recent advancements in deep learning and machine learning to make early and accurate predictions of AD and distinguishing it from other neurodegenerative diseases. 
 
 In this paper, we propose a non-invasive method of predicting and distinguishing AD from other neurodegenerative diseases. We have trained ML models using the speech (audio) transcripts of individuals with different neurodegenerative diseases.
This paper also studies the impact of speech features on model predictions. More formally, the paper studies which of the myriad speech features are the most relevant in making accurate predictions for the trained models (and hence in diagnosis of AD). Intuitively, here's how speech plays a critical part in the diagnosis of AD: early stage AD patients commonly face difficulty recalling words, or finding the right vocabulary. Verbal fillers, such as “um” are common and slower speech occurs. During moderate stages, slurring, stammering, repetition and use of incorrect words or phrases are common. In the final stages of AD, patients may lose the ability to form coherent thoughts and speech. Individuals frequently repeat phrases they hear from others. When speech occurs (if at all), it is more often than not incoherent or illogical. Individuals sing, babble or say words unrelated to the situation and conversation \cite{c:22}.
 While the speech problems mentioned above are commonly observed in AD patients, they can also provide indications of other neurodegenerative diseases that can cause speech disorders, such as mild cognitive impairment (MCI), vascular dementia and aphasia. Thus, speech can give a great insight in predicting \& distinguishing AD from other neurodegenerative diseases. But the next obvious question that arises is that which features or which parts of speech can actually indicate presence of AD. Through our research, we plan to analyse a person's speech in an attempt to understand which factors of speech can be used to diagnose Alzheimer's disease.

The remaining part of the paper is organized as follows. Section 2 talks about the related work done before in this field and how it inspired us in our study. Section 3 describes the dataset and it's basic preprocessing. Section 4  talks about the evaluation metrics used for judging the performance of our models. Section 5 contains the exact technical details of the project, presents the various experiments we performed along with the results we obtained and is further divided into two subsections, one each for model-wise and feature analysis. Finally, section 6 wraps up the paper with conclusion and future work.

\section{Related Works}

A relatively small subset of studies on Alzheimer's disease attempt to quantify the impairments in connected speech using computational techniques. Primitive works on the subject include the study conducted by Bucks et al.\cite{c:11} who studied a linear discriminant analysis of spontaneous speech  from  8  AD  participants  and  16  healthy  controls. They considered eight linguistic features, including part-of-speech  (POS)  tag  frequencies  and  measures of lexical  diversity, and obtained a cross-validation accuracy of 87.5\%. Lin  et  al.\cite{c:1} conducted  their  study on  Dem@care  dataset  by  using  spectrogram features in speech data for AD patients  and  training logistic regression to get the best results (outperforming all other models), other models being Decision Tree, linear SVC and MLP. C. Thomas et al.\cite{c:4} relied on character n-gram based techniques on ACADIE Dataset, showing ML classification techniques to be viable on AD dataset and classifying different stages of Dementia of Alzheimer type (DAT). Their work depicted that classifying dementia into multiple categories loses accuracy, but it is highly distinctive in distinguishing control from DAT patients (95\%).

Studies performed on the DementiaBank dataset (that we have used in our work) include that of Wankerl et al. \cite{c:12}, who employed an n-gram based model on Pitt Corpus (DementiaBank), and using the Equal Error Rate (EER) as the classification threshold they achieved an accuracy of 77.1\%, as well as Mini-Mental State Examination (MMSE) scores of the test subjects. Their work discerns that while little can be said about the speech of healthy control group, a higher correlation could be found when considering the demented speakers. Yancheva, M. et al.\cite{c:19} worked on the DementiaBank dataset to predict MMSE scores  based  on  linguistic  features  of  AD and  related  dementia. They used a bivariate dynamic Bayes net to represent longitudinal progression of observed linguistic features and MMSE scores over time. Their focus on individuals with more longitudinal samples improved the mean absolute error value, suggesting the importance of longitudinal data collection. Fraser et al.\cite{c:3} performed work on the transcripts and acoustic labels of audio from DementiaBank data and used logistic regression to classify. They achieved a maximum accuracy of 81.92\% in distinguishing between AD and controls. Their work also included scatter plots between different kinds of speech factors and their inter-relationships. In one of the most recent works on this database, in 2018, Karlekar  et  al. \cite{c:13} applied neural networks  on  utterances from four speech tasks in Pitt corpus and obtained the best accuracy of 91.1\%. However, they broke each speech transcript into individual utterances and treated those utterances as independent samples. In this way, the model tends to lose the ability to check thematic coherence in the speech. Another popular work on the DementiaBank dataset includes that of Chien et al. \cite{c:5} They designed bidirectional RNNs and used AUROC as the metric of evaluation. The focus of their work was on real world applications for distinguishing AD from non-AD patients. Chen et al.\cite{c:6} also worked on the DementiaBank dataset, training an attentive neural network  (Att-CNN+Att-BiGRU architecture) and achieved the best accuracy (97.42\%) among all other models. 

None of the previously done works attempt to perform a qualitative analysis of speech in predicting and classifying AD. On top of that, most of those works have attempted to classify AD with only 2 or 3 other neurodegenerative diseases. Our study adds to the previously done work in several ways:
\begin{itemize}
    \item Training and comparing performances of multiclass classifiers for distinguishing between 6 classes: probable AD, possible AD, MCI, memory impairments, vascular dementia and control and binary classifiers for distinguishing between Probable AD and control.
    
    \item Performing a qualitative study on the impact of various speech features in predicting AD and classifying it from other neurodegenerative diseases.
\end{itemize} 
\vspace{1mm}
\section{Dataset}

For our work in this paper, we have used the DementiaBank's Pitt Corpus dataset which contains transcripts of conversations with people, many of whom had different forms of dementia, acquired in CHAT Format (Codes for the Human Analysis of Transcripts).

We chose this dataset as lots of work has been done on it in the past, and hence multiple performance benchmarks were available to us for comparison \cite{c:6}. It is also one of the most widely used datasets in Alzheimer's related research that has been performed prior to our work. The dataset contains speech transcripts of patients coming from different parts of the world, having upto 9 different types of neurodegenerative diseases (as can be seen in Table~\ref{table:datadistribution}) and has been manually annotated, the specifics of which can be found in the works of Becker et al. \cite{c:16} and Lopez et al. \cite{c:17}. The precise criteria used to categorize subjects into the different classes present in DementiaBank's Pitt Corpus dataset, can be found in the work of McKhann et al. \cite{c:14}. The huge diversity in the dataset (a testament to the rigour of the experiment with which the transcripts of conversations were collected), indicates that it is a good reflection of the real world data, and hence the models trained on it would generalize well and would probably perform well in real life.

The transcripts were parsed using the CLAN software \cite{c:10} to extract features specific to our needs using the following commands :
\begin{itemize}
    \item IPSYN for syntactic complexity.
    \item EVAL for word-type ratios, grammatical analysis and count of utterances.
    \item FLUCALC for fluency and pauses.
\end{itemize}

Table~\ref{table:datadistribution} shows the class-wise composition of the dataset.
\begin{table}[hbt!]
\centering 
\caption{Data Distribution} 
\begin{tabular}{c c} 
\hline 
Group & No. of Samples \\ [0.5ex] 
\hline 

ProbableAD & 762 \\ 
Control & 243 \\
MCI & 162 \\
PossibleAD & 68 \\
Vascular & 20 \\ 
Memory & 12 \\ 
Other & 4 \\ 
Dementia & 1 \\
Uncategorised & 1 \\ [1ex] 
\end{tabular}
\label{table:datadistribution} 
\end{table}

\begin{table}[hbt!]
\centering 
\caption{Features Retained After Preprocessing}
\begin{tabular}{c c}
\hline 
Category & Feature  \\ [0.5ex] 
\hline 
\\
IPSYN & N, V, Q, S  
\\\\[1ex] 
EVAL &  \makecell{Age, Sex, Group, Duration\_(sec),\\
 MLU\_Utts, MLU\_Morphemes,\\
 FREQ\_TTR, Words\_Min, 
 Verbs\_Utt, \\\%\_Word\_Errors,
 Utt\_Errors, density,\\ \%\_Nouns,
 \%\_Plurals, \%\_Verbs,\\ \%\_Aux,
 \%\_Mod, \%\_3S,  \%\_13S,\\
 \%\_PAST, \%\_PASTP, \%\_PRESP,\\
 \%\_prep, \%\_adj,
 \%\_adv, \%\_conj, \\\%\_det,
 \%\_pro, noun\_verb}
 \\\\[1ex]
 FLUCALC & \makecell{
  retracing, repetition, mor\_Utts,\\ 
  mor\_syllables, syllables\_min,\\
  \%\_Prolongation, Mean\_RU, \\
  \%\_Phonological\_fragment, \\
\%\_Phrase\_repetitions,\\ 
\%\_Word\_revisions, \\
\%\_Phrase\_revisions,\\
\%\_Pauses, \%\_Filled\_pauses,\\
\%\_TD, SLD\_Ratio,\\
Content\_words\_ratio,\\
Function\_words\_ratio}\\
 [1ex] 
\hline 
\end{tabular}
\label{table:featuretable} 
\end{table}

%Insert Correlation Matrix
\begin{figure}[ht]
%\begin{center}
\includegraphics[width= 0.475\textwidth]{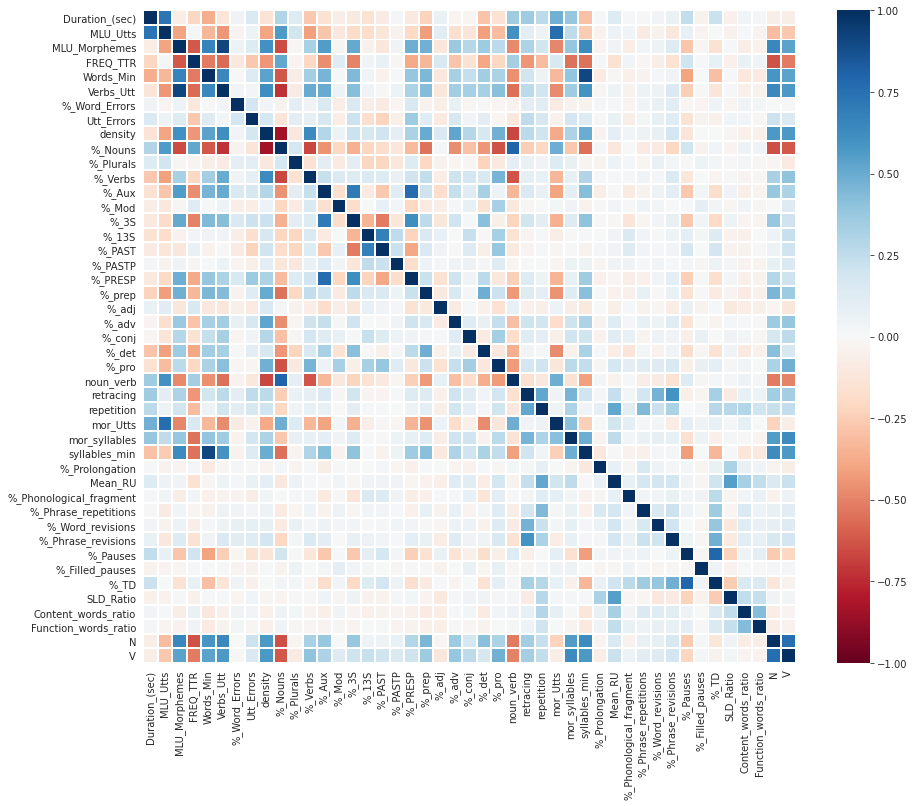}
\caption{Correlation Matrix}
\label{fig:arch1}
%\end{center}
\end{figure}

Out of the 9 classes present in the dataset, we have removed the classes: Other, Dementia and Uncategorised,  and used the remaining 6 for our work, for the simple reason that they had too little data points for any model to generalize well on them. 
Initially there were a total of 108 features in the data. Selected features were retained by observing the correlation matrix shown in Fig.~\ref{fig:arch1}. In any pair of features having a correlation value of more than 0.8, one of them was removed. Eventually, 50 features were retained from a total of 108 features in the beginning. Table~\ref{table:featuretable} shows the features we retained for our work after preprocessing and cleaning the data.

\section{Evaluation Metrics}

Evaluating the machine learning models is an essential part of any machine learning or deep learning project. This forms the basis on which changes are made to model architectures in an attempt to attain the best results and to compare their performances. To understand how well the model was performing, we calculated the following 4 evaluation metrics:
\begin{itemize}
\setlength\itemsep{1em}
    \item Recall: It is the fraction of correctly predicted positive samples out of all positive samples in the dataset.
    % ANOTHER DEF:It is the ratio of positive predictions that were correct to the predictions that ideally should have been positive.
    \begin{displaymath}
    Recall  = \frac{True\ Positives}{True\ Positives + False\ Negatives}
    \end{displaymath}
    \item Precision: It is the fraction of positive predictions that were correct out of the total positive predictions made by the model.\\
    \begin{displaymath}
    Precision  = \frac{True\ Positives}{True\ Positives + False\ Positives}
    \end{displaymath}
    \item Accuracy: The most common performance metric, it is simply the ratio of the number of correct predictions to the total number predictions.\\
    \begin{displaymath}
    Accuracy  = \frac{True\ Positives + True\ Negatives}{Total}
    \end{displaymath}
    \item F1-Score: Formally, F1-score is the harmonic mean between precision and recall. It essentially tries to find the perfect balance between precision and recall. 
    \\
    \begin{displaymath}
    F1\mbox{-}Score  = \frac{2\times Precision \times Recall}{Precision + Recall}
    \end{displaymath}
\end{itemize}

For our study, wherein early detection is essential (and is in fact one of the main goals), false negatives would mean that a potential AD case was missed by the classifier. If the classifier predicts false positives, the person can rule out the disease when they take a professional’s opinion to start the treatment, but if the classifier does not recognise an actual positive case (false negative), the loss can be significantly larger in real life. Hence the training was done, keeping in mind to minimize the number of false negatives (and hence the recall). That was one of the reasons why we chose to go with 4 different evaluation metrics as opposed to having only accuracy.

% Table of Model Hyperparameters
\begin{table*}[hbt!]
\centering 
\caption{Model Hyperparameters} 
\begin{tabular}{c c c} 
\hline 
Model & Hyperparameters(Binary) & Hyperparameters(multiclass)\\ [0.5ex] 

\hline 
\\
XGB Classifier & 
\makecell {
colsample\_bytree: 0.65,
max\_depth: 2, \\
n\_estimators: 90
}&
\makecell {
colsample\_bytree: 0.55, 
max\_depth: 2, \\
n\_estimators: 90
}\\
[1ex] \\ [1ex]
Support Vector Classifier & 
\makecell {
C: 11.20, 
kernel: rbf \\
}&
\makecell {
C: 1.55, 
kernel: rbf \\
}\\
[1ex] \\ [1ex]
Decision Tree & 
\makecell {
min\_depth: 9, 
min\_samples\_split: 0.005 \\
}&
\makecell {
min\_depth: 5, 
min\_samples\_split: 0.039 \\
}\\
[1ex] \\ [1ex]
Random Forest & 
\makecell {
max\_depth: 8, 
min\_samples\_split: 0.022, \\
n\_estimators:240\\
}&
\makecell {
max\_depth: 16, 
min\_samples\_split: 0.010, \\
n\_estimators:20\\
}\\
[1ex] \\ [1ex]
Neural Networks & 
\makecell {
No. of layers: 5\\
Neurons (in increasing depth): 64, 128, 256, 64, 2\\
Training time: 60 epochs, 
Optimizer: Adam, \\
Learning rate: 0.004 (constant)\\
} &
\makecell {
No. of layers: 8\\
Neurons (in increasing depth): 64, 128, 256, 256, 256, 128, 64, 6\\
Training time: 40 epochs, 
Optimizer: Adam, \\
Learning rate: 0.004  initially, decreased to 0.8 times\\
its current value every 17 epochs\\
}\\

\hline
\end{tabular}
\label{table:hyperparameters}
\end{table*}

\section{Experiments, Results \& Analysis}

We trained multiple classifiers to make predictions and distinguish AD from other neurodegenerative diseases. Both binary and multiclass models were trained, with the former just distinguishing between Probable AD and Control, whereas the multiclass classification distinguishes among all 6 classes present after cleaning the dataset: ProbableAD, Control, MCI, PossibleAD, Vascular, and Memory. As mentioned previously, the remaining 3 groups (classes) present in Table~\ref{table:datadistribution}, weren't used because of their extremely small sample size and lack of proper definition. The following models were trained: KNN, SVC, Decision Tree, Random Forest, XGB classifier and neural networks. Table~\ref{table:hyperparameters} contains the fine-tuned hyperparameters obtained using grid search for each of these models (other than neural networks). Apart from the hyperparameters mentioned for neural nets, they also included dropout (mean value of 0.3) and batch normalization after every layer (barring the output layer), to counter overfitting and to improve the generalizing ability of the model.
For each of the aforementioned models, we have calculated and compared the accuracy, the weighted values of precision, recall and F1-score, i.e, the weighted mean of class-wise measures according to the number of samples present. 
Taking the weighted mean of these metrics helps in giving equal importance to each class, even for classes with a small sample size when performing multiclass classification.

Furthermore, confusion matrices were plotted model-wise that allowed us to make in-depth inferences about the performances of the models we had trained after all the hyperparameter tuning. Secondly, a thorough speech feature analysis was done, to find which features in speech are most important for a model while making predictions. The following two subsections contain the discussion about the results of model-wise as well as feature analysis.

\subsection{Model-wise Analysis}
\begin{table*}[t!]
\centering
\caption{Classifier Performance} 
\begin{tabular}{c c c c c c c c c} 
\hline
Classifier & \multicolumn{2}{c}{Accuracy} & \multicolumn{2}{c}{Weighted Precision} & \multicolumn{2}{c}{Weighted Recall} &   \multicolumn{2}{c}{Weighted F1} \\ [0.5ex] 
\hline
\\
  & \emph{All Groups} & \emph{Binary} & \emph{All Groups} & \emph{Binary} & \emph{All Groups} & \emph{Binary} & \emph{All Groups} & \emph{Binary} \\
   [1ex]

XGB Classifier & 75.59\% & 89.55\% & 0.76 &  0.89 & 0.76 & 0.90 & 0.73 & 0.89\\[1ex]

\makecell{Support Vector\\Classifier} & 70.47\% & 90.04\% & 0.64 &  0.90 & 0.70 & 0.90 & 0.65 & 0.90 \\[1ex]
Decision Tree & 68.11\% & 85.57\% & 0.69 &  0.86 & 0.68 & 0.86 & 0.64 & 0.86\\[1ex]

Neural Networks & 76.44\% & 92.05\% & 0.72 &  0.93 & 0.76 & 0.93 & 0.72 & 0.91\\[1ex]

\makecell{K-Nearest\\Neighbours} & 72.05\% & 88.56\% & 0.68 & 0.91 & 0.72 & 0.89 & 0.68 & 0.89\\[1ex]

Random Forest & 70.47\% & 86.07\% & 0.63 &  0.85 & 0.70 & 0.86 & 0.64 & 0.85\\[1ex]

\hline
\end{tabular}
\label{table:classifierperformance} 
\end{table*}
From Table~\ref{table:classifierperformance}, we can observe that neural networks have outperformed the other models in terms of accuracy for both binary and multiclass classification. The table also contains the weighted values of precision, recall and F1-score attained by all models for both binary and multiclass classification.
While precision and recall seem to be somewhat inconclusive, in terms of F1-score, it would be safe to say that for both types of classifications, neural networks performed almost as good as the other models, if not better. In other words, neural network model has been able to better optimize both precision and recall as compared to other models.

The following inferences were made from the model-wise confusion matrices plotted for the predictions made on the test data:
\begin{itemize}
    \item In multiclass classification, some classes (mostly Memory and Vascular, but at times PossibleAD as well) were not predicted at all for the test data. The most probable reason for this is the presence of very few samples for these classes as can be seen in  Table~\ref{table:datadistribution}. A small sample size for any particular class causes the models to overfit as it is unable to learn the underlying patterns thus fails to generalize on real world data. %However, for Possible AD, which was often classified as Probable AD, %[Insert difference between Probable and Possible] can be rightfully be considered, since we are trying to evaluate the possibility of Alzheimers’ being present through speech.
    \item Another notable observation from our results is that in all multiclass classifiers that we trained, a significant fraction of participants with MCI were being categorised as either ProbableAD or Control. 
    This seems to align with the idea that MCI can be viewed as a very early stage of AD \cite{c:21}.
    Research has shown that individuals with MCI have a significantly higher risk of developing AD within a few years, compared to people with normal cognitive function. Research surrounding MCI offers other potential paths to earlier diagnosis of AD \cite{c:8}.
    
    Our experimentation indicates that MCI cannot be accurately distinguished from AD using only linguistic features. For better differentiation, we recommend using automatic speech analyses for the assessment of MCI and early-stage AD, as shown in the works of König A et al.\cite{c:18}
    
    \item Participants with memory impairments were also quite commonly categorised as cases of ProbableAD. This might be due to the fact that AD also affects memory. However, since less data was available for the group “Memory”, a conclusion cannot be made just from this observation and may require further evidence.
\end{itemize}

In multiclass classification, the models usually failed to perform well for the groups that had very small number of samples in the dataset. This is, of course a pitfall of using a quite unbalanced (albeit diverse) dataset. In the future, we plan to work on a larger and more evenly balanced dataset in order to draw any strong conclusions about such groups.

\subsection{Feature Analysis} Studying feature importance in classifying Alzheimer's from other neurodegenerative diseases was one of the primary goals of our work. As can be seen in Fig.~\ref{fig:arch}, for both binary as well as multiclass classification, we have plotted the feature importance graphs for each classifier, except KNN, SVC and neural networks, in order to study which speech features were really important for making predictions for our models.
The feature importance was not plotted for the three  aforementioned classifiers for the following reasons:
\begin{itemize}
    \item KNN does not provide a prediction for the importance or coefficients of variables since it gives equal importance to each of the ‘k' neighbouring data points.
    
    \item SVC has been trained using the Radial Basis Function (RBF) kernel. Since this kernel transforms the features and classification is done using the transformed features, the importance or coefficients cannot be traced back to the original features.
    
    \item A neural network is not a linear model and hence, the feature importance cannot be determined in terms of the original input features.
\end{itemize}

\begin{figure*}[ht]
\begin{center}
\includegraphics[width= 1\textwidth]{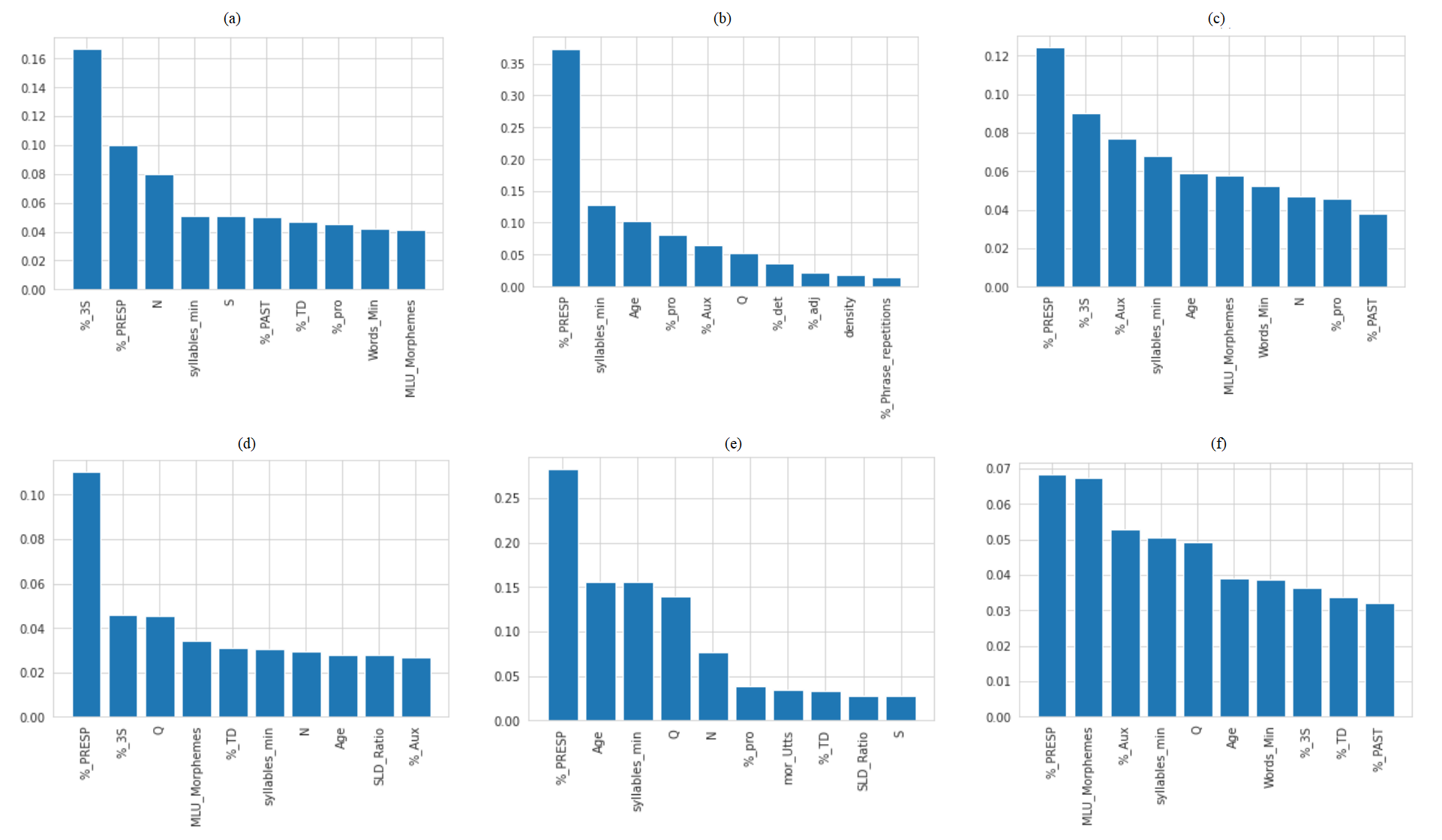}
\caption{Feature Importance Graphs : (a) Binary XGB Classifier, (b) Binary Decision Tree Classifier, (c) Binary Random Forest Classifier, (d) Multiclass XGB Classifier, (e) Multiclass Decision Tree Classifier, (f) Multiclass Random Forest Classifier}
\label{fig:arch}
\end{center}
\end{figure*}

 The set of graphs shown in Fig.~\ref{fig:arch} display the importance of features model wise. The y-axis denotes how important a particular feature is (calculated as the decrease in node impurity weighted by the probability of reaching that node), and the x-axis denotes the 10 most important features (in decreasing order of importance, from left to right) relevant in making predictions for a particular model. The higher the value of feature importance, the more important the feature.

The observations from the graphs in Fig.~\ref{fig:arch} suggest that ‘\%\_PRESP’ (present participle) and ‘\%\_3S’ (3rd person present tense markers) are the most relevant features for all models and for both the tasks, in general. 

Present participle (or \%\_PRESP) is a form of verb that uses ‘-ing' with the base of the word. Most of the time, it performs the function of an adjective, though at times, it also works as a verb or a subject in construction. It is important to note that this is not the same as a gerund, where the verb is serving as a noun. The 3rd person present tense marker (or \%\_3S) is a verb phrase exclusively characterized by the use of the -s inflection with the third person singular present tense \cite{c:15}. The following examples elaborate the same -
\begin{itemize}
\item Present Participle : “He is coughing” (instead of “Coughing exhausts him”, where ‘coughing’ is a gerund)

\item 3rd Person present tense marker : “She cycles” (instead of “She cycled” or “Donna cycles”).
\end{itemize}

Our observations suggest that speech of patients with AD consists of present participle excluding gerunds and includes the use of present tense markers with 3rd person reference.
In all, the features, present participle and 3rd person present tense markers are the most pertinent among all features in the distinguishing AD from other neurodegenerative diseases.

\section{Conclusion} 

Inspired by the recent developments in the field of neuroscience and deep learning, we in this paper trained various models (both binary and multiclass) to predict whether a person has AD and to distinguish it from other neurodegenerative diseases by using speech transcripts, on a publicly available dataset. An exhaustive comparative study between the performance of models, was conducted by plotting confusion matrices and using 4 different performance metrics and neural networks were found to perform the best in both types of classification tasks. 

In the paper, we also managed to show that there exists a strong correlation between speech of a person and whether he/she has AD or not. This was established by a comprehensive feature analysis that was done, which allowed us to make reasonable inferences about the relevance of certain features in helping a model make predictions.
We plan to extend our work on a larger and a much more balanced dataset in the future, with which we can study biases and experiment more heavily in order to improve the generalizability of our models. Development of artificial data via GANs or similar techniques might be a possible solution if enough base data can be acquired. Encouraged by our findings, we hope that our work would serve as a good reference point for other research projects aiming to make the diagnosis of AD easier and its early detection possible.

\bibliographystyle{ieeetr}
% \balance
\bibliography{bibliography.bib}
\end{document}